\documentclass{svmult}
    
\usepackage{mathptmx}  
\usepackage{helvet} 
\usepackage{courier}
\usepackage{amsmath,amsfonts}
\usepackage{graphicx}
\usepackage[bottom]{footmisc}          
   
\usepackage[table]{xcolor}    
\usepackage{mathptmx}         
\usepackage{helvet}             
\usepackage{courier}             
\usepackage{type1cm}             
\usepackage{makeidx}             
\usepackage{graphicx}          
\usepackage{multicol}          
\usepackage[bottom]{footmisc} 
    
\usepackage{epsfig}     
\usepackage{psfrag,rotating}     
\usepackage{amssymb,floatflt,enumerate} 
\usepackage{amsmath,amscd,psfrag,leqno} 
\usepackage[mathscr]{eucal}  
\usepackage{shadethm}           
\usepackage{graphicx,subfigure}   
\usepackage{overpic,contour}   
\contourlength{0.3mm}  
 
\def\Beweisende{\square}            
\def\BewEnde{\hfill{\Beweisende}}

\def\phm{{\hphantom{-}}}

\def\phi{\varphi}

\def\RR{{\mathbb R}}

\def\dach#1{\widehat{#1}}

\def\Vkt#1{{\mathbf #1}}

\newcommand{\mVkt}[1]{\dach{\Vkt #1}}

\newcommand{\dotVkt}[1]{\dot{\Vkt #1}}

\newcommand{\go}[1]{{\sf #1}}

\makeindex
\begin{document}

\title*{Kinematically Redundant Octahedral Motion Platform for Virtual Reality Simulations}
\author{G. Nawratil and A. Rasoulzadeh}
\authorrunning{G. Nawratil and A. Rasoulzadeh}
%\institute{$^1$IRCCyN, CNRS, France, \email{Philippe.Wenger@irccyn.ec-nantes.fr} \\
%$^1$University of Minho, Portugal, \email{pflores@dem.uminho.pt}}
\institute{
  Institute of Discrete Mathematics and Geometry, Vienna University of Technology, Austria, 
  \email{\{nawratil,rasoulzadeh\}@geometrie.tuwien.ac.at}}

\maketitle

\abstract{We propose a novel design of a parallel manipulator of Stewart Gough type for 
virtual reality application of single individuals; i.e. an 
omni-directional treadmill  
is mounted on the motion platform in order to improve VR immersion
by giving feedback to the human body.
For this purpose we modify the well-known octahedral manipulator in a way that it has one degree of kinematical redundancy; 
namely an equiform reconfigurability of the base. 
The instantaneous kinematics and singularities of this mechanism 
are studied, where especially "unavoidable singularities" are characterized. 
These are poses of the motion platform, which can only be 
realized by singular configurations of the mechanism despite its kinematic redundancy. 
}


\section{Introduction}

The geometry of a Stewart-Gough (SG) platform is given by the six base anchor points $\go M_i$ with coordinates 
$\Vkt M_i:=(A_i,B_i,C_i)^T$ with respect to the fixed frame and by the six platform anchor points $\go m_i$ 
with coordinates $\Vkt m_i:=(a_i,b_i,c_i)^T$ with respect to the moving frame (for $i=1,\ldots ,6$). 
Each pair $(\go M_i,\go m_i)$ of corresponding anchor points of the fixed body (base) and the 
moving body (platform) is connected by an S\underline{P}S-leg, where only 
the prismatic joint (\underline{P}) is active and the spherical joints (S) are passive. 
The distance between the centers of the two S-joints of the $i$th leg is denoted by $r_i$. 
Note that for a SG platform, $(\go M_i,\go m_i)\neq (\go M_j,\go m_j)$ holds for pairwise distinct 
$i,j\in\left\{1,\ldots ,6\right\}$. 
Moreover if two base points $\go M_i$ and $\go M_j$ (resp.\ platform points $\go m_i$ and $\go m_j$) coincide, then 
we just write $\go M_{i,j}$ (resp.\ $\go m_{i,j}$).

A certain drawback of such 6-degree of freedom (dof) robotic platforms  
is the limitation of their singularity-free workspace. 
A promising approach for overcoming this problem is redundancy,  where basically two types can 
be distinguished:
\begin{enumerate}[(a)]
\item
{\bf Actuation redundancy:} 
One possibility is to add a 7th S\underline{P}S-leg to the manipulator (e.g.\ \cite{dasgupta,zhang,cao}), but 
it should be noted that there exist dangerous locations for the attachments $\go m_7$ and $\go M_7$ (cf.\ \cite{hmm}), 
which do not result in a reduction of singularities. 
Clearly more than one S\underline{P}S-leg can be attached, but we have to keep in mind that 
every extra leg may yields a restriction of the workspace due to leg constraints and leg interference \cite{dasgupta}.
Moreover \cite{schreiber} pointed out that 
this so-called branch redundancy\footnote{A further possibility to 
imply actuation redundancy in a SG platform is to replace passive joints by active ones. No 
attempts to this so-called in-branch redundancy are known to the authors.} 
causes internal loads, whose control increases considerably the 
complexity and costs. 
\item
{\bf Kinematic redundancy:} 
This can be achieved by attaching active joints in a way that the geometry of the platform or base (or both) can 
be modified. Therefore they are also known as {\it reconfigurable manipulators}\footnote{The modules in 
a variable geometry truss (VGT) can also be seen as reconfigurable manipulators (cf.\ \cite[Fig.\ 9]{merlet_redundant}). 
In this context see also \cite{stoughton}, where a VGT is combined with a SG structure.} of SG type. 
In \cite{bande} a design is suggested, which has six redundant dofs (3 in the platform and 3 in the base). 
The weak points of this so-called {\it dodekapod} are that on the one hand additional mass/inertia (motor of active joints)
is added to the platform and that on the other hand passive prismatic joints are used, which are difficult to implement in practice 
(cf.\  \cite{schreiber}). 
Not faced with these points of criticism are the designs suggested in \cite{wang,kotlarski}, where 
one base point is traced on a circle (realized by an extra \underline{R}-joint \cite{wang}) 
or a straight line (realized by an extra \underline{P}-joint \cite{kotlarski}). 

We complete this review by noting a different approach to kinematic redundancy; namely  
the usage of {\it parallel redundant legs} introduced in \cite{schreiber}.

\end{enumerate}
Another advantage of approach (b) over (a) beside the already mentioned ones above is that 
(a) increases the workspace only by removing singularities, but (b) additionally 
enlarges the workspace itself due to the reconfigurability of the anchor points.  
Further differences between both concepts are pointed out at the end of Section \ref{sec:unavoidable}.

\subsection{Aim and basic concept}
Within the Doctoral College "Computational Design" of the Vienna University of Technology, 
we are interested amongst other things in the optimization of motion platforms of SG type for 
virtual reality (VR) application of single individuals. In detail, we want to mount an 
omni-directional treadmill onto the motion platform (cf.\ Fig.\ \ref{fig2}) in order to improve VR immersion
by giving feedback to the human body (e.g.\ simulating slopes, earthquakes, \ldots).
There are different locomotion systems commercially available; e.g.\ 
Cyberith Virtualizer (\begin{small}{\tt cyberith.com}\end{small}), %\cite{virtualizer}, 
Omni Virtuix (\begin{small}{\tt www.virtuix.com}\end{small}), % \cite{omni}, 
SpaceWalkerVR (\begin{small}{\tt spacewalkervr.com}\end{small}), %\cite{swvr}, 
Kat Walk (\begin{small}{\tt www.katvr.com}\end{small})  %\cite{kat} 
and
Wizdish ROVR (\begin{small}{\tt www.wizdish.com}\end{small}). %\cite{wd}. 

Due to safety reasons we prefer a system where the individuals are fixed inside a hip-ring   
by belts and straps. Thus we are only left with the Omni Virtuix and the Cyberith Virtualizer. 
We favor the latter system as it has one more dof, which allows to kneel/crouch/sit down or even 
to jump up, and a flat base plate with a suitable friction coefficient for simulating slopes. 
Contrary, the Omni Virtuix's base has a very low friction (too slippery for slope simulations) 
and a concave shape.

\begin{figure}[t!]
\begin{center} 
\begin{overpic}
    [height=34mm]{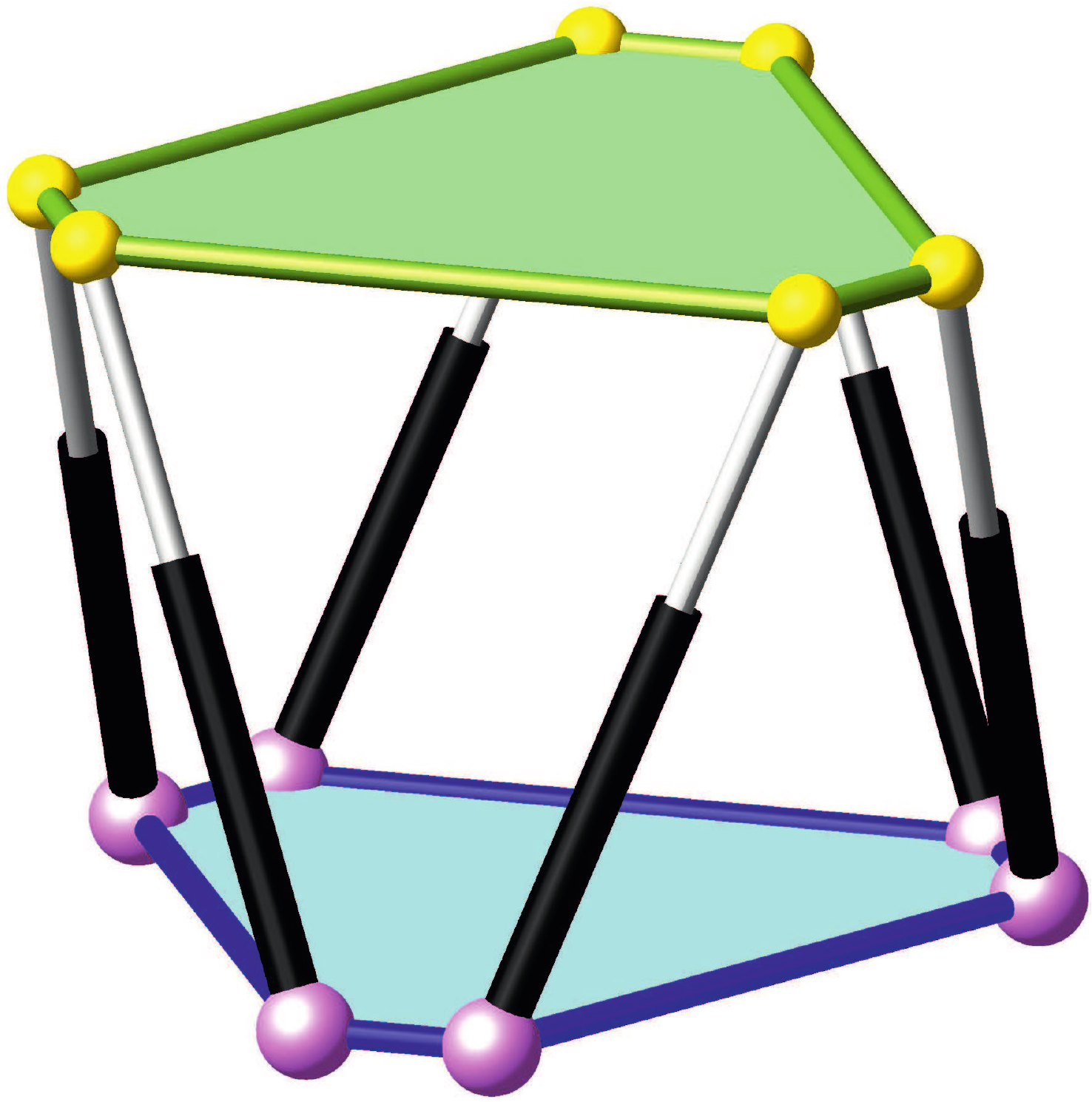}
  \end{overpic} 
\qquad \qquad \qquad
 \begin{overpic}
    [height=36mm]{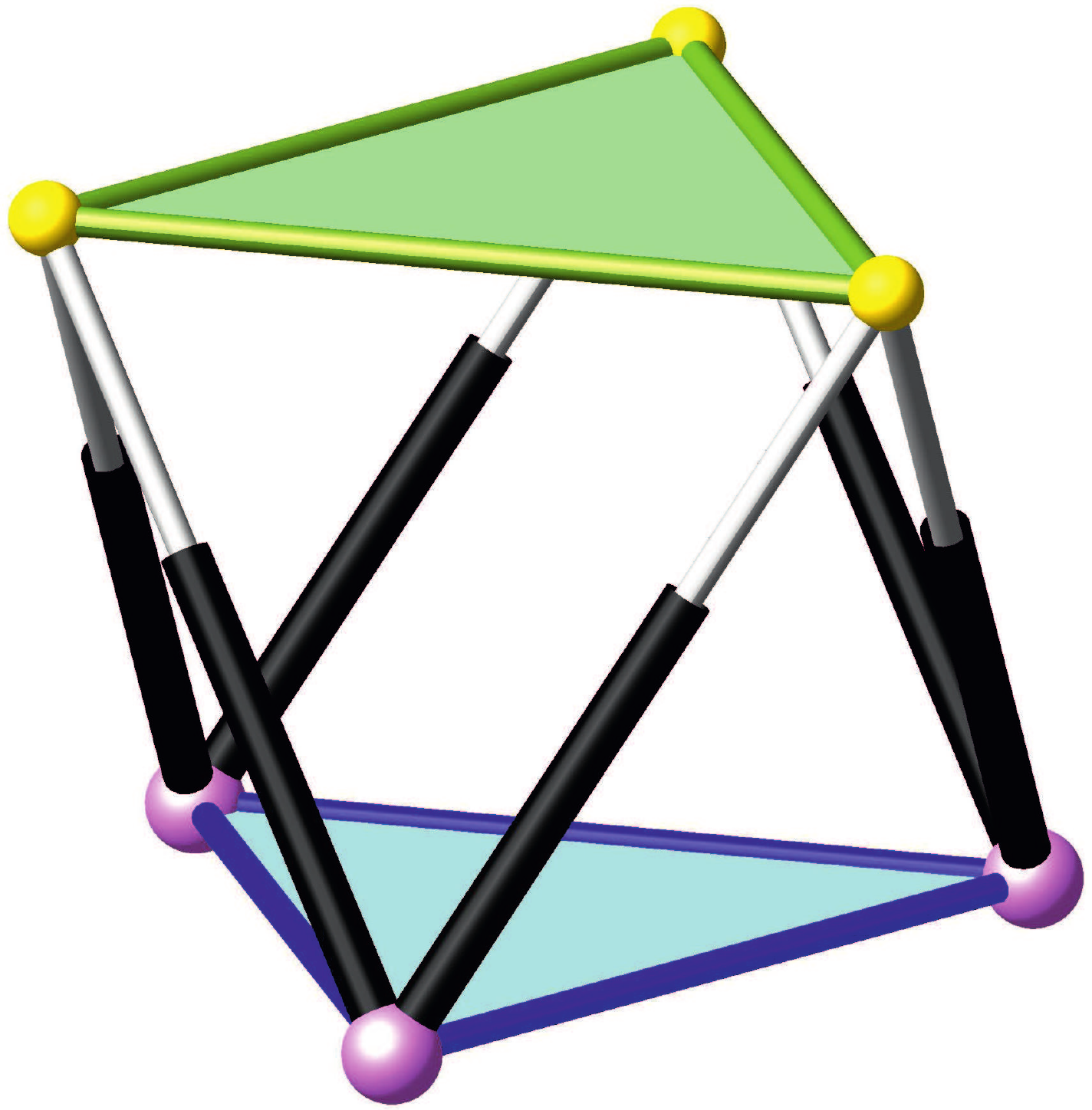}
\begin{small}
\put(-21,86){$\go m_5=\go m_6$}
\put(-3,0){$\go M_6=\go M_1$}
\put(-17,19){$\go M_4=\go M_5$}
\put(85,9){$\go M_2=\go M_3$}
\put(85,73){$\go m_1=\go m_2$}
\put(68,94.5){$\go m_3=\go m_4$}
\end{small}         
  \end{overpic} 
\end{center} 
\caption{
(Left) SG manipulator with planar semi-hexagonal base and platform. 
(Right) Octahedral SG manipulator, which is also known as MSSM (Minimal Simplified Symmetric Manipulator) in the literature (e.g.\ \cite{merlet_buch}).
This parallel robot is characterized by:
$\go m_1=\go m_2$, 
$\go m_3=\go m_4$,  
$\go m_5=\go m_6$,  
$\go M_2=\go M_3$,  
$\go M_4=\go M_5$, 
$\go M_6=\go M_1$. 
Therefore all attachments are double spherical joints, which form a triangle in the platform and the base. 
Moreover we assume that both triangles are equilateral. 
}
\label{fig1}
\end{figure}
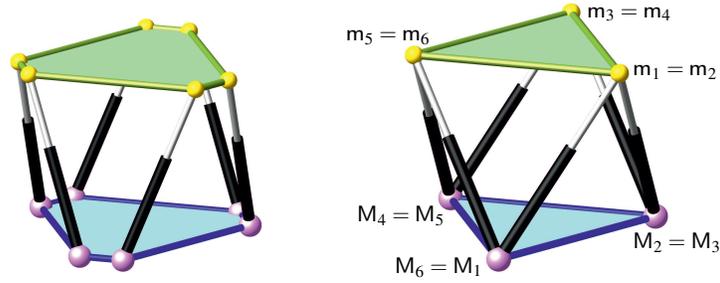

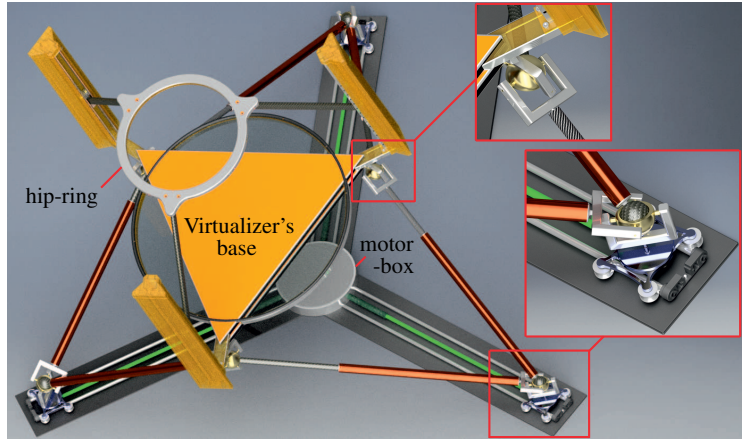
\begin{figure}[t!]
\begin{center} 
 \begin{overpic}
    [height=58mm]{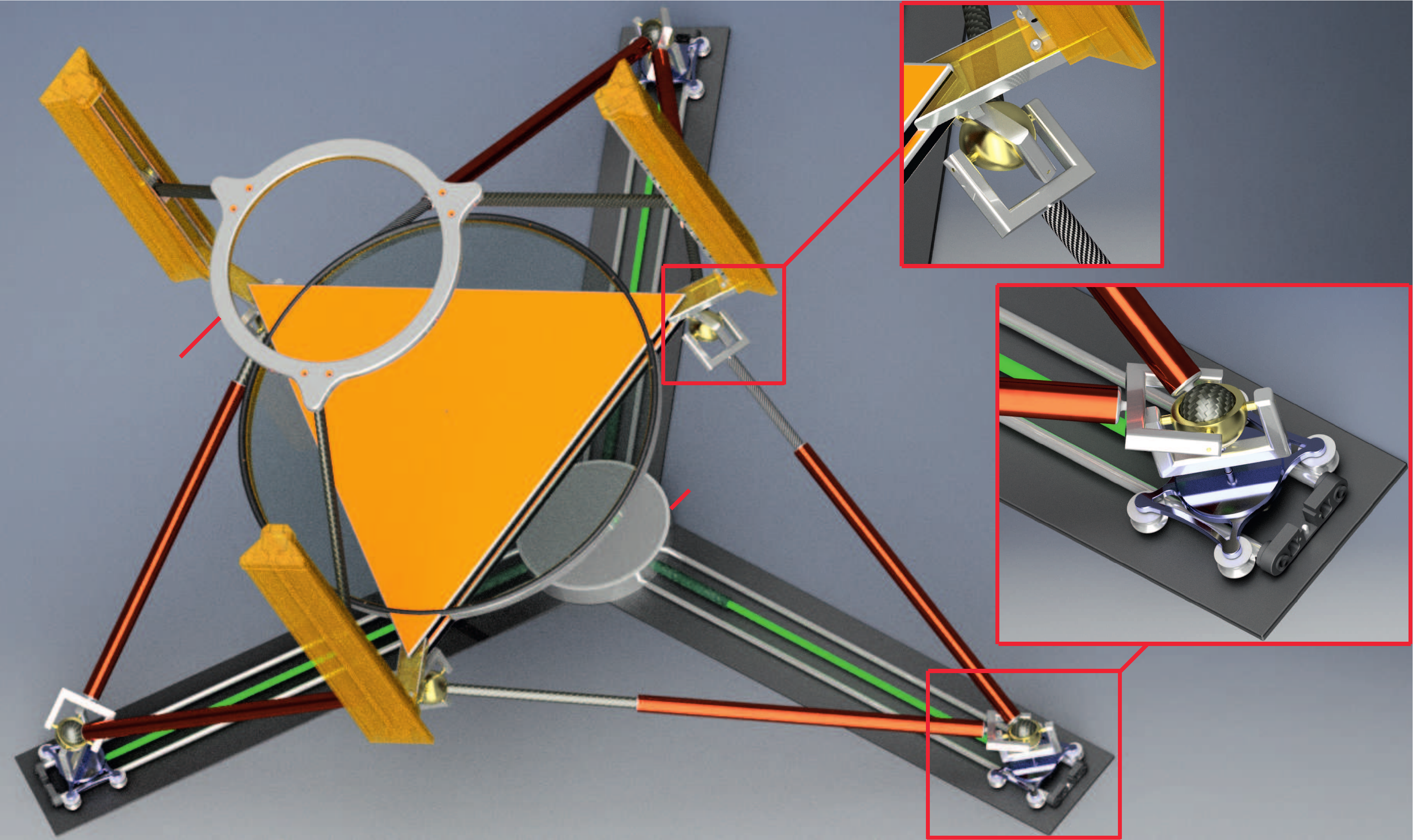}
\begin{small}
\put(2.5,32){hip-ring}
\put(24,28){Virtualizer's}
\put(28.5,25){base}
\put(48,25.5){motor}
\put(49.8,22.5){-box}
\end{small}       
  \end{overpic} 
\end{center} 
\caption{Octahedral SG manipulator with a 1-dof kinematic redundancy (equiform reconfigurability of the 
base) and a Virtualizer mounted on top. In practice one S-joint of an S\underline{P}S-leg is replaced by 
an universal joint (U), thus all double S-joints can be replaced by concentric SU-joints (blow-ups).
}
\label{fig2}
\end{figure}

All currently offered 6-dof motion simulators of SG type have a planar semi-hexagonal base 
and platform (see Fig.\ \ref{fig1} (left)). This design is preferred over the so-called octahedral SG 
manipulator (see Fig.\ \ref{fig1} (right)) as it avoids double S-joints, which 
have a lot of disadvantages regarding costs, stiffness and joint-range. 
Nevertheless we focus within this  
{\it concept study} on the octahedral design 
(as many other academic studies e.g.\  
\cite{stoughton}, 
\cite[Section 4]{wang}, 
\cite{merlet_buch,fichter,hunt,quality,downing,hamdoon,jiang}, 
\ldots
done in the past), due to its 
kinematic simplicity. But this application-alien abstraction can also be brought back to 
real-world solutions as pointed out in Section \ref{outlook}. 

Due to the arguments given above we  want to use concept (b) 
of kinematic redundancy in order to increase the singularity-free workspace 
of the octahedral manipulator. Having our practical application in mind, 
the resulting reconfigurable manipulator should still be of kinematic simplicity 
(the symmetry of the manipulator should not be destroyed) and cheap 
; i.e.\ we want to get by with 
standard components (less or even no expensive custom products). 

Taking all these requests under consideration, we came up with the 
design illustrated in Fig.\ \ref{fig2}. 
We use three additional active \underline{P}-joints (green), which are 
driven simultaneously by only one motor mounted at the 
center of the base. Therefore the base triangle remains equilateral 
during the reconfiguration process. 
The proposed manipulator 
can be seen as a combination of a SG hexapod and a so-called 
hexaglide (e.g.\ \cite{bhebsacker}). As the linear sliding of the points $\go M_{i,i+1}$ 
along the lines $\go g_{i,i+1}$ can be realized by parallel guiding rails, 
the whole structure can still be built very stiff with a high
load-carrying capacity. Moreover the complete additional equipment for the 
performance of the reconfiguration (motor, rails, \ldots) is placed on the resting base 
such that the legs and platform are not stressed by additional mass/inertia.

\begin{remark}\label{rem:1}
Let us assume that the three additional prismatic joints are passive instead of active (but still coupled). 
Moreover we lock the six \underline{P}-joints of the legs. Then the base has 
in each pose of the manipulator at least a 1-dimensional self-motion with respect to the platform, 
due to the equiform reconfigurability of the base. These relative motions (of the base with respect to the platform) 
are studied in \cite{hamdoon}. \hfill $\diamond$
\end{remark}
\newpage

\section{Instantaneous kinematics and singularities}
The $i$th leg of the 7-dof reconfigurable manipulator fulfills the basic relation 
$r_i(\tau)^2=\|\Vkt n_i(\tau)-\Vkt M_i(\tau)\|^2$, 
where $\Vkt n_i$ is the coordinate vector of $\go m_i$ with respect to the fixed system.  
Differentiation with respect to the time $\tau$ yields:
\begin{equation}
r_i\dot r_i = (\Vkt n_i-\Vkt M_i)(\dotVkt n_i -\dotVkt M_i) \quad \Rightarrow \quad
\dot r_i=\Vkt l_i(\dotVkt n_i -\dotVkt M_i),
\end{equation}
where $\Vkt l_i$ is the unit-vector along the $i$th leg (pointing from the base point to the platform point). 
As the velocity of $\Vkt n_i$ can be expressed in terms of the instantaneous screw  
$(\Vkt q,\mVkt q)$  of the platform with respect to the fixed system
by $\dotVkt n=\mVkt q+\Vkt q\times \Vkt n_i$ we get:
\begin{equation}
\dot r_i + \Vkt l_i\dotVkt M_i = \Vkt l_i \mVkt q + \Vkt l_i(\Vkt q\times \Vkt n_i) \quad \Rightarrow \quad
\dot r_i + \dot g\Vkt g_i\Vkt l_i = \Vkt l_i \mVkt q +\mVkt l_i\Vkt q,
\end{equation}
where $\Vkt g_i$ is the unit-vector along $\go g_i$, $\dot g$ the velocity along $\go g_i$
and $\mVkt l_i$ the moment vector of the $i$th leg. Note that $(\Vkt l_i,\mVkt l_i)$ 
are the spear coordinates of the $i$th leg.
Thus we end up with: 
\begin{equation}\label{relationv}
\begin{pmatrix}
\dot r_1 \\
\vdots \\
\dot r_6
\end{pmatrix}
+\dot g
\begin{pmatrix}
\Vkt g_1\Vkt l_1 \\
\vdots \\
\Vkt g_6\Vkt l_6
\end{pmatrix}
=
\Vkt J
\begin{pmatrix}
\Vkt q \\
\mVkt q
\end{pmatrix}\quad
\text{with}\quad
\Vkt J:=
\begin{pmatrix}
\,\,\mVkt l_1^T & \phm\Vkt l_1^T \\
\,\,\vdots & \phm\vdots \\
\,\,\mVkt l_6^T & \phm\Vkt l_6^T
\end{pmatrix}.
\end{equation} 
This equation relates the velocities $\dot r_1, \ldots, \dot r_6, \dot g$ of the 
active \underline{P}-joints with the instantaneous screw $(\Vkt q,\mVkt q)$ of the platform. 
If we fix all prismatic joints the left side of  Eq.\ (\ref{relationv}) equals the zero vector. 
Then an infinitesimal motion of the platform is only possible if the Jacobian matrix $\Vkt J$ of the 
octahedral manipulator is singular; which is well studied problem 
(e.g.\ \cite{hunt,downing} or \cite[pages 202--204]{merlet_buch}).

\subsection{Unavoidable singularities}\label{sec:unavoidable}

\begin{definition}
If for a given pose of the platform the corresponding configuration of the mechanism is singular 
for all kinematically redundant dofs, we call 
this singular pose of the platform an "unavoidable singularity". 
\end{definition}

To the best knowledge of the authors there are no studies on "unavoidable singularities" 
in the literature, with the following two exceptions:
\begin{enumerate}[$\bullet$]
\item
In \cite[Section V(D)]{schreiber} it is stated that for the octahedral manipulator with three 
redundant legs {\it "any Cartesian pose of the moving platform can be reached with a non-singular 
configuration of the mechanism".}  
Moreover, it is "{\it conjectured that this result can be extended to 
mechanisms whose spherical joints on the platform do not coincide by pairs"}, but 
this is disproved in Appendix A. 
\item
For the kinematically redundant SG manipulator of \cite{wang}, there exist up to 40 positions of the 
platform causing an unavoidable singularity for a given orientation (cf.\ \cite[Section 4.4.2]{wang}). 
Some special orientations causing a higher-dimensional set of unavoidable singularities were 
studied in \cite[Section 4.4.3]{wang}. 
\end{enumerate}
For the proposed mechanism of the paper at hand, we can chose an appropriate scaling factor and 
coordinate systems in the fixed and moving space in a way that:
\begin{align}
\Vkt m_{1,2}&=(1,0,0)^T, &\quad
\Vkt m_{3,4}&=(-\tfrac{1}{2},\tfrac{\sqrt{3}}{2},0)^T, &\quad
\Vkt m_{5,6}&=(-\tfrac{1}{2},-\tfrac{\sqrt{3}}{2},0)^T, \\
\Vkt M_{4,5}&=g(-1,0,0)^T, &\quad
\Vkt M_{6,1}&=g(\tfrac{1}{2},-\tfrac{\sqrt{3}}{2},0)^T, &\quad
\Vkt M_{2,3}&=g(\tfrac{1}{2},\tfrac{\sqrt{3}}{2},0)^T,
\end{align}
hold, where $g>0$ is the reconfiguration variable ($=$ radius of the circumcircle of the equilateral base triangle).
Based on this coordinatization we can compute the entries of $\Vkt J$ by $(\Vkt l_i:\mVkt l_i):=
(\Vkt R\Vkt m_i + \Vkt s-N\Vkt M_i:\Vkt M_i\times \Vkt l_i)$ with the rotation matrix 
\begin{equation*}
\Vkt R = 
\begin{pmatrix} 
e_0^2+e_1^2-e_2^2-e_3^2 & 2(e_1e_2-e_0e_3) & 2(e_1e_3+e_0e_2)  \\
2(e_1e_2+e_0e_3) & e_0^2-e_1^2+e_2^2-e_3^2 & 2(e_2e_3-e_0e_1)  \\
2(e_1e_3-e_0e_2) & 2(e_2e_3+e_0e_1) & e_0^2-e_1^2-e_2^2+e_3^2
\end{pmatrix},
\end{equation*}  
$N=e_0^2+e_1^2+e_2^2+e_3^2$ and the translation vector $\Vkt s=(x,y,z)^T$. Note that $(e_0:e_1:e_2:e_3)\neq (0:\ldots :0)$ 
are the so-called Euler parameters. 

Then the singularity condition $\Sigma$: $\det\Vkt J=0$  has the following structure: 
\begin{equation}
\Sigma:\quad 83\sqrt{3}Ng^3(Q[12]N^2g^2+L[56]Ng+A_1[29]A_2[8])=0,
\end{equation}
where the number in the brackets gives the number of terms. 
Note that $Q$ and $A_2$ depend linearly on $x,y,z$ and that $L$ and $A_1$ 
are polynomials in $z^2,xz,yz,x,y,z$. 

\begin{theorem}\label{thm1}
The unavoidable singularities of the 1-dof kinematic redundant octahedral manipulator of SG type studied in the paper at hand 
are listed in the following table containing the positions of the unavoidable singularities in the 3rd column for the 
orientations given in the 2nd column and the 4th column gives the dimension of the singularity set:

\begin{small}
\begin{center}
\begin{tabular}{|c|c|c|c|c|}
\hline 
row & orientation & position & dim   \\ \hline \hline 
1 & $e_1=e_2=0$ & $z=0$ & 3     \\ \hline % field of lines 
2 & $e_1=e_2=0$, $e_0=\pm e_3$  & arbitrary & 3  \\ \hline % Fichter singularity
3 & $e_0=e_3=0$ & arbitrary & 4  \\ \hline % regular linear line complex, moving platform upside down
4 & $e_0=\pm e_3$, $e_2=\mp e_1$ & $z=2e_1e_3$ & 3  \\ \hline %singular linear line complex 
5 & $e_0=\pm e_3$, $e_2=\mp e_1$ & $z = -\frac{e_3(2e_3^2+2e_1^2\pm y)}{e_1}$, $x=0$ & 2  \\ \hline % regular linear line complex
6 & $e_0=\pm e_3$, $e_2=\mp e_1$ & $z = -4e_1e_3$, $y = \pm 2(e_1^2-e_3^2)$ & 2  \\ \hline   % regular linear line complex
7 & $e_0=e_3$, $e_1= (2\pm \sqrt{3})e_2$ & $z=\frac{4e_2e_3}{1\mp \sqrt{3}}$ & 3  \\ \hline % singular linear line complex 
8 & $e_0=e_3$, $e_1= (2\pm \sqrt{3})e_2$ & $x=\frac{16e_2^2\pm8e_2^2\sqrt{3}+y-4e_3^2}{\pm\sqrt{3}}$, $z=\frac{8e_2e_3}{\pm \sqrt{3}-1}$ & 2  \\ \hline
9 & $e_0=e_3$, $e_1= (2\pm\sqrt{3})e_2$ & $x=\mp y\sqrt{3}$, $z = \frac{2e_3(e_3^2+4e_2^2\pm 2e_2^2\sqrt{3}-y)}{e_2(1\pm \sqrt{3})}$ & 2  \\ \hline
10 & $e_0=-e_3$, $e_1= (-2\pm \sqrt{3})e_2$ & $z=\frac{4e_2e_3}{\mp \sqrt{3}-1}$ & 3  \\ \hline % singular linear line complex 
11 & $e_0=-e_3$, $e_1= (-2\pm \sqrt{3})e_2$ & $x=\frac{16e_2^2\mp 8e_2^2\sqrt{3}-y-4e_3^2}{\mp\sqrt{3}}$, $z=\frac{8e_2e_3}{1 \pm \sqrt{3}}$ & 2  \\ \hline
12 & $e_0=-e_3$, $e_1= (-2\pm\sqrt{3})e_2$ & $x=\mp y\sqrt{3}$, $z = \frac{2e_3(e_3^2+4e_2^2\mp 2e_2^2\sqrt{3}+y)}{e_2(\pm \sqrt{3}-1)}$ & 2  \\ \hline
13 & $e_0=e_2=0$  & $z=e_1e_3$ & 3  \\ \hline
14 & $e_1=e_3=0$  & $z=-e_0e_2$ & 3  \\ \hline
15 & $e_0=e_1=0$  & $y=z=0$ & 2  \\ \hline
16 & $e_2=e_3=0$  & $y=z=0$ & 2  \\ \hline
17 & $e_0e_1+e_2e_3=0$ & $z$ of Eq.\ (\ref{coord:z})  & 4 \\ \hline
18 & $e_2^2e_3^2-3e_2^2e_0^2+8e_2e_1e_0e_3-3e_1^2e_3^2+e_1^2e_0^2=0$ & $z$ of Eq.\ (\ref{coord:z}) & 4 \\ \hline
19 & $e_0=\frac{e_2e_3}{e_1}$  & $y=2e_2\frac{e_1^2+e_3^2}{e_1}$, $z = \frac{e_3(e_1^4-6e_1^2e_2^2+e_2^4)}{e_1(e_1^2-e_2^2)}$ & 3 \\ \hline
20 & $e_0=-\frac{e_1e_3}{e_2}$ & $y=2e_1\frac{e_3^2-e_2^2}{e_2}$, $z = -4e_1e_3$ & 3 \\ \hline
21 & arbitrary & $z$ of Eq.\ (\ref{coord:z}), $x$ of  Eq.\ (\ref{coord:x1}), $y$ of  Eq.\ (\ref{coord:y1}) & 3  \\ \hline
22 & arbitrary & $z$ of Eq.\ (\ref{coord:z}), $x$ of  Eq.\ (\ref{coord:x2}), $y$ of  Eq.\ (\ref{coord:y2}) & 3  \\ \hline
\end{tabular}
\end{center}
\end{small}
with
\begin{small}
\begin{align}\label{coord:z}
z &= \frac{(e_0^2+e_3^2)(e_1^3e_3-3e_1e_2^2e_3-3e_1^2e_2e_0+e_2^3e_0)}{(e_3^2-e_0^2)(e_2^2+e_1^2)}, \\
\label{coord:x1}
x &= \frac{(e_0^2+e_3^2)(e_3^2e_1^2-e_3^2e_2^2-4e_1e_2e_0e_3-e_1^4+6e_2^2e_1^2-e_2^4-e_0^2e_1^2+e_0^2e_2^2)}{2(e_2^2+e_1^2)(e_0^2-e_3^2)}, \\
\label{coord:y1}
y &= \frac{(e_3^2+e_0^2)(e_1e_2e_3^2+e_1^2e_3e_0-e_2^2e_3e_0+2e_1^3e_2-2e_1e_2^3-e_1e_2e_0^2)}{(e_2^2+e_1^2)(e_3^2-e_0^2)}, \\
\label{coord:x2}
x &= \frac{e_3^4e_1^2-e_3^4e_2^2+3e_3^2e_1^2e_2^2-e_3^2e_2^4+2e_3e_0e_1^3e_2+2e_3e_0e_1e_2^3-e_1^4e_0^2+3e_1^2e_2^2e_0^2-e_0^4e_1^2+e_0^4e_2^2} 
{(e_2^2+e_1^2)(e_0^2-e_3^2)}, \\
\label{coord:y2}
y &= \frac{2e_3^4e_1e_2+3e_3^2e_1^3e_2+e_3e_2^4e_0-e_3^2e_1e_2^3-e_3e_0e_1^4-2e_0^4e_1e_2+e_0^2e_1^3e_2-3e_0^2e_1e_2^3}{(e_2^2+e_1^2)(e_3^2-e_0^2)}.
\end{align}
\end{small}
\end{theorem}

\vfill
\begin{proof}
For an unavoidable singularity the conditions  $Q=L=A_1A_2=0$
have to hold. This system of equations can be solved explicitly\footnote{In contrast to \cite{wang}, 
where the unavoidable singularities are obtained as the numeric solution of an univariate polynomial of degree 40.} 
for $x,y,z$. The detailed case study yielding the above table is presented in Appendix B.  Note that 
for a given orientation the positions are (unions of) affine subspaces of $\RR^3$.
\hfill $\BewEnde$
\end{proof}
\newpage

As the unavoidable singularities are low-dimensional varieties (cf.\ Theorem \ref{thm1}) 
in the $6$-dimensional workspace, they can always be bypassed 
during path-planning. But this does not imply the existence of a singularity-free path as 
the singularity variety is a hypersurface in 
the mechanism's configuration space.\footnote{For our mechanism this is a $7$-dimensional space, as 
a mechanism configuration is  given by the pose of the platform (6 dofs) and by the size of the base triangle (1 dof).} 
Due to topological reasons two points of the configuration space can be separated by this hypersurface. 

In contrast, actuation redundancy reduces the dimension of the singularity variety 
by the number of redundant dofs (cf.\ \cite[Section 3]{dasgupta}) and 
therefore a singularity can always be avoided during path-planning.

\begin{figure}[t!]
\begin{center} 
\begin{overpic}
    [height=40mm]{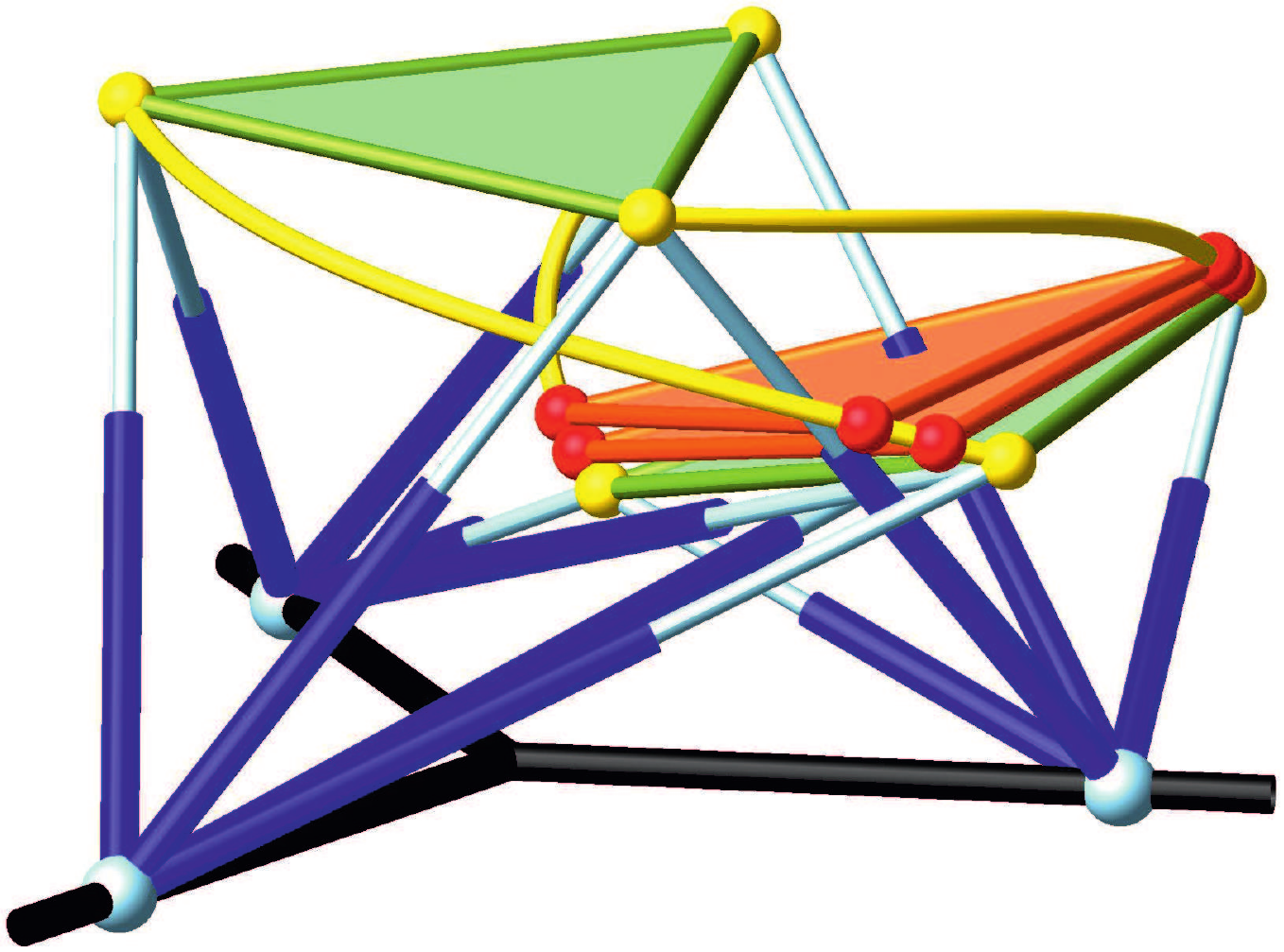}
		\begin{small}
\put(13,71){start}
\put(81.5,35){end}
\end{small}         
  \end{overpic} 
	\qquad\qquad
 \begin{overpic}
    [height=40mm]{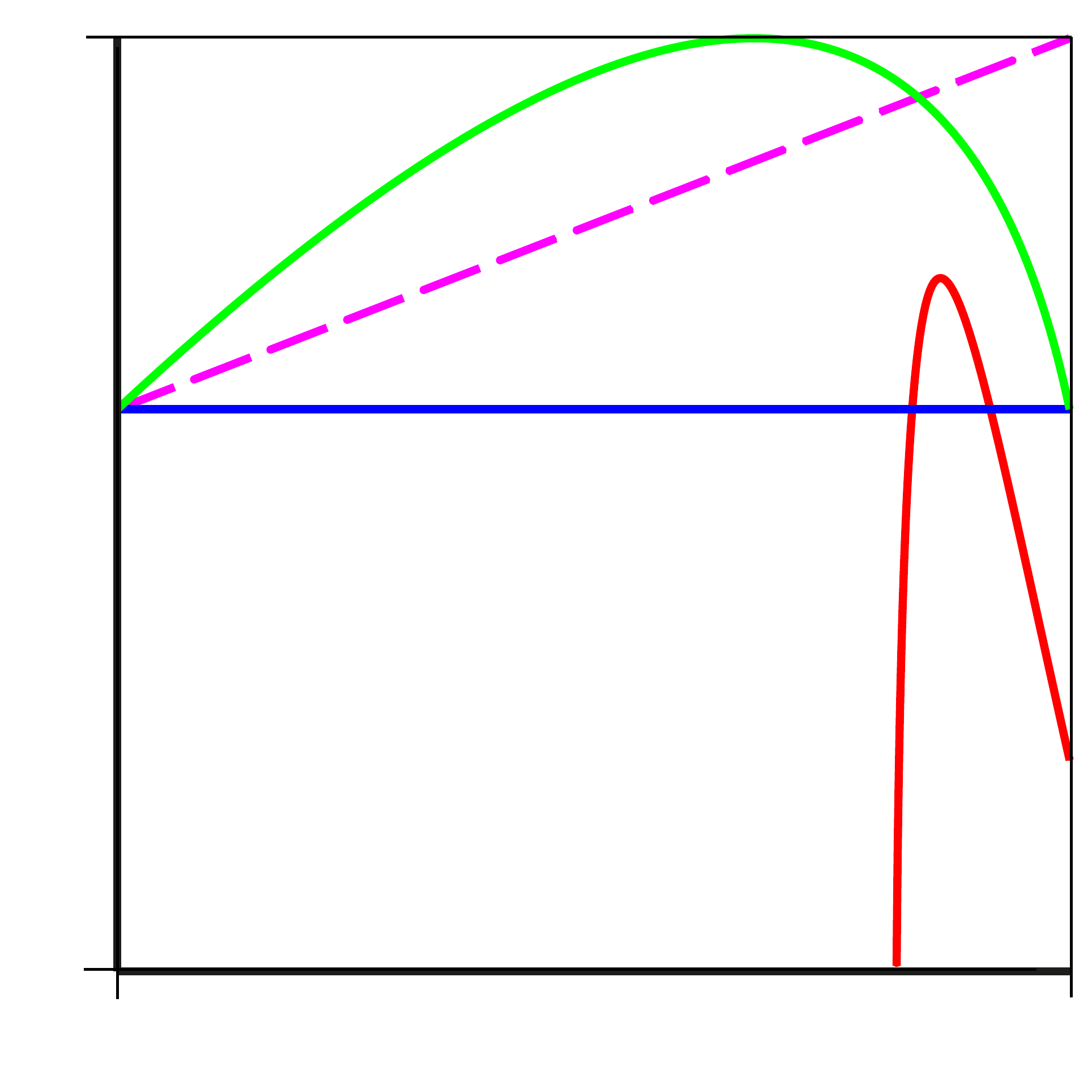}
		\begin{small}
\put(-5.5,10){min}
\put(-6.5,95){max}
\put(2.5,50){g}
\put(5.5,1){start}
\put(27.5,1){motion parameter}
\put(92,1){end}
\begin{turn}{90}
\put(16,-79){singularities}
\end{turn}
\end{small}         
  \end{overpic} 
\bigskip\newline	
\begin{overpic}
    [height=40mm]{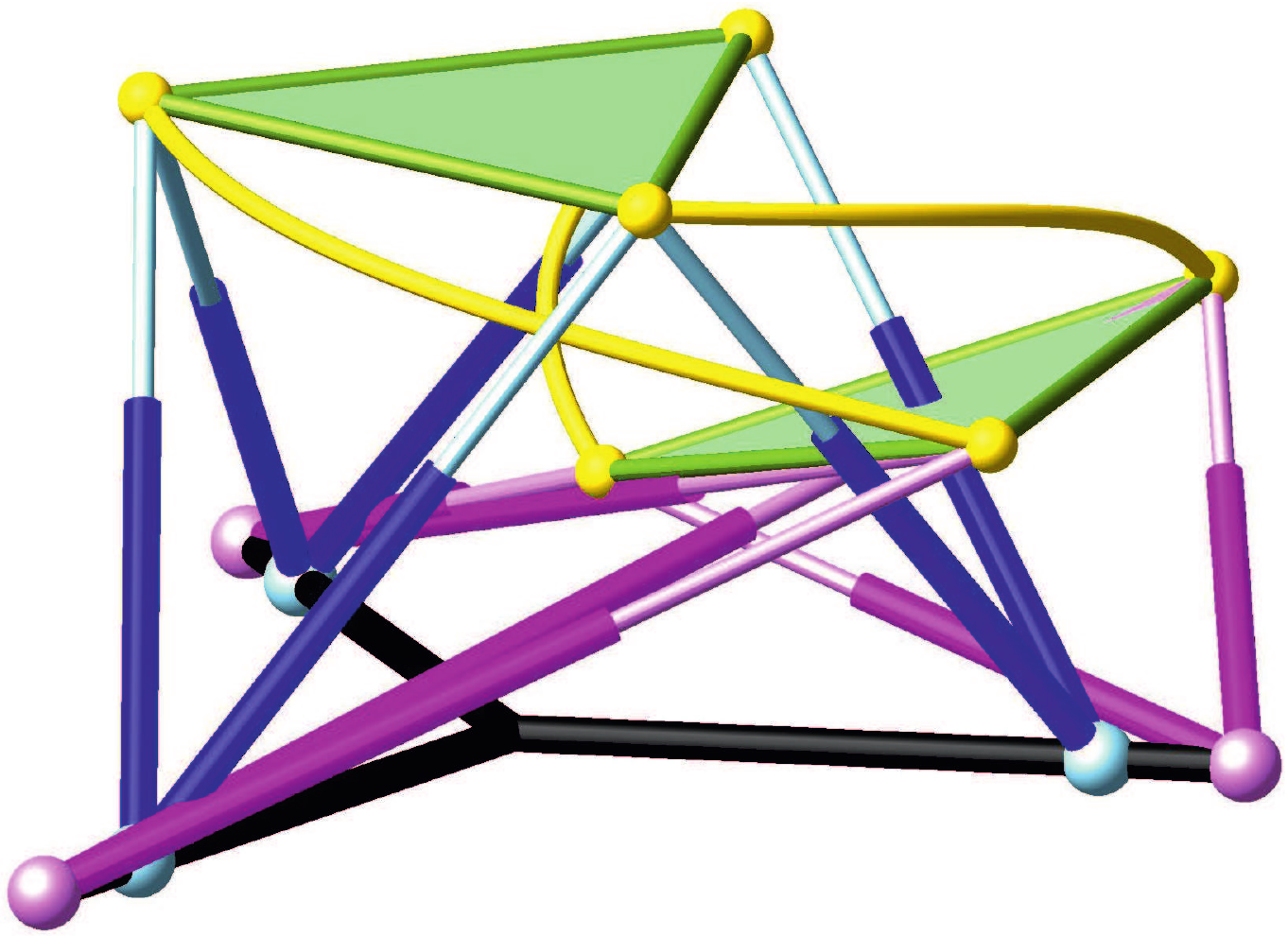}
		\begin{small}
\put(14,69){start}
\put(81,35){end}
\end{small}         
  \end{overpic}
\hfill
\begin{overpic}
    [height=39mm]{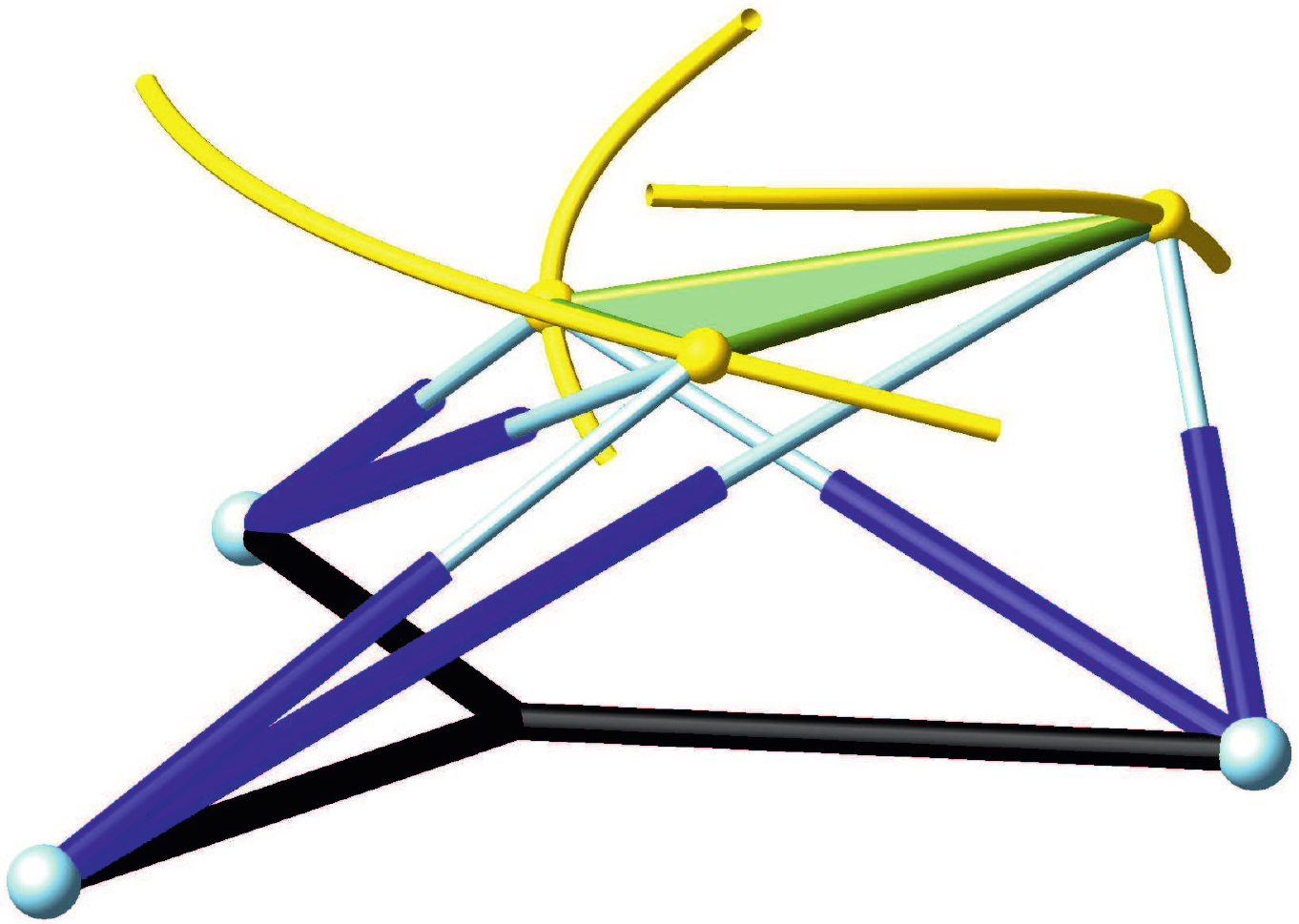}       
  \end{overpic}
\end{center} 
\caption{Singularity avoidance by kinematical redundancy: In the upper left picture the given motion of the 
platform between the start- and end-pose is displayed. Without reconfiguration of the base the 
manipulator will pass two singular configurations, which are highlighted in red. They correspond to 
the two intersection points of the blue horizontal line and the red {\it singularity curve} in the diagram in the 
upper right corner. One can avoid these singularities by performing a reconfiguration of the base indicated by the 
dashed line in the diagram, but then there is a collision of legs in the end-pose, which is illustrated in the lower left picture. 
The green parabola in the diagram corresponds to a solution, which is free of leg-interference and singularities. 
The robot-configuration of this solution with $g=max$ is displayed in the lower right picture.
An animation can be downloaded from: {\tt http://www.geometrie.tuwien.ac.at/nawratil/reconfiguration.gif} 
}
\label{figbsp}
\end{figure}
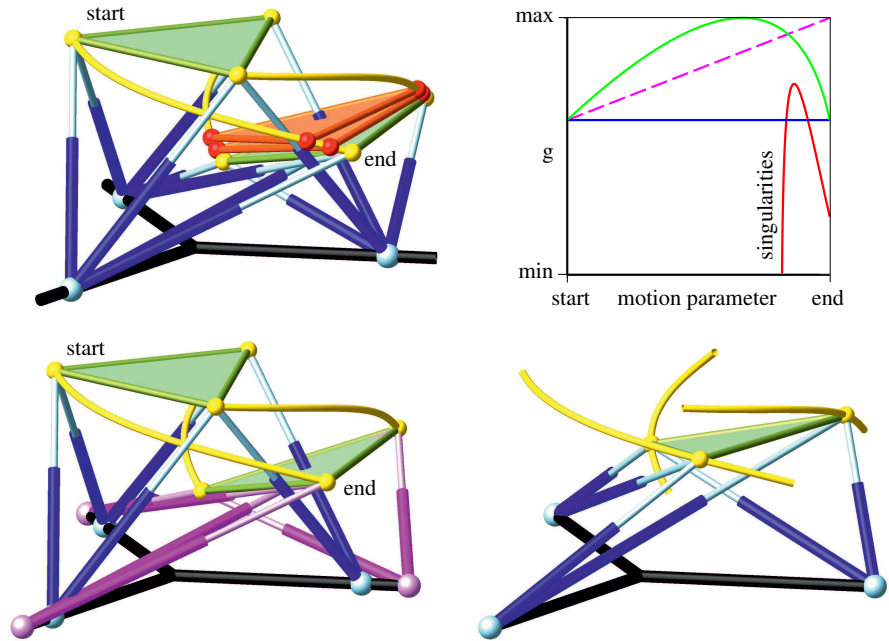

But in the case of a given path of the platform, 
kinematic redundancy is superior to actuation redundancy, as kinematically
redundant dofs can be used to avoid singularities (if possible) and to increase 
the performance of the manipulator during the prescribed motion (cf.\ Fig.\ \ref{figbsp}). 
The study of the optimal reconfiguration of the base of the proposed mechanism 
during a given motion is dedicated to future research. 

\begin{remark}
Note that a so-called modular parallel robot \cite{merlet_modular} also changes its geometry with respect to 
a given end-effector motion, but there is no reconfiguration of the robot's base/platform {\it during} the 
motion of the end-effector. \hfill $\diamond$ 
\end{remark}

\section{Conclusion}\label{outlook}

This paper contains a {\it concept study} of an octahedral SG manipulator with one degree of 
kinematic redundancy (equiform reconfigurability of the base) for VR simulations of single individuals (see Fig.\ \ref{fig2}). 
In a side-note we also proposed SU-joints (see Fig.\ \ref{fig2}) as an alternative to the problematic double S-joints 
within the octahedral structure. Clearly a deeper study of these SU-joints (regarding stiffness, joint-range, dynamic behavior,  \ldots) 
has to be done in order to judge their applicability. 
Another possibility is the substitution of all double S-joints by double U-joints (e.g. see 
\cite[Fig.\ 16]{fichter} and \cite[Section 4]{quality}) and to replace the \underline{P}-joints 
by cylindrical joints, with a passive rotation and an active translation component. 

\begin{figure}[t!]
\begin{center} 
\begin{overpic}
    [height=42mm]{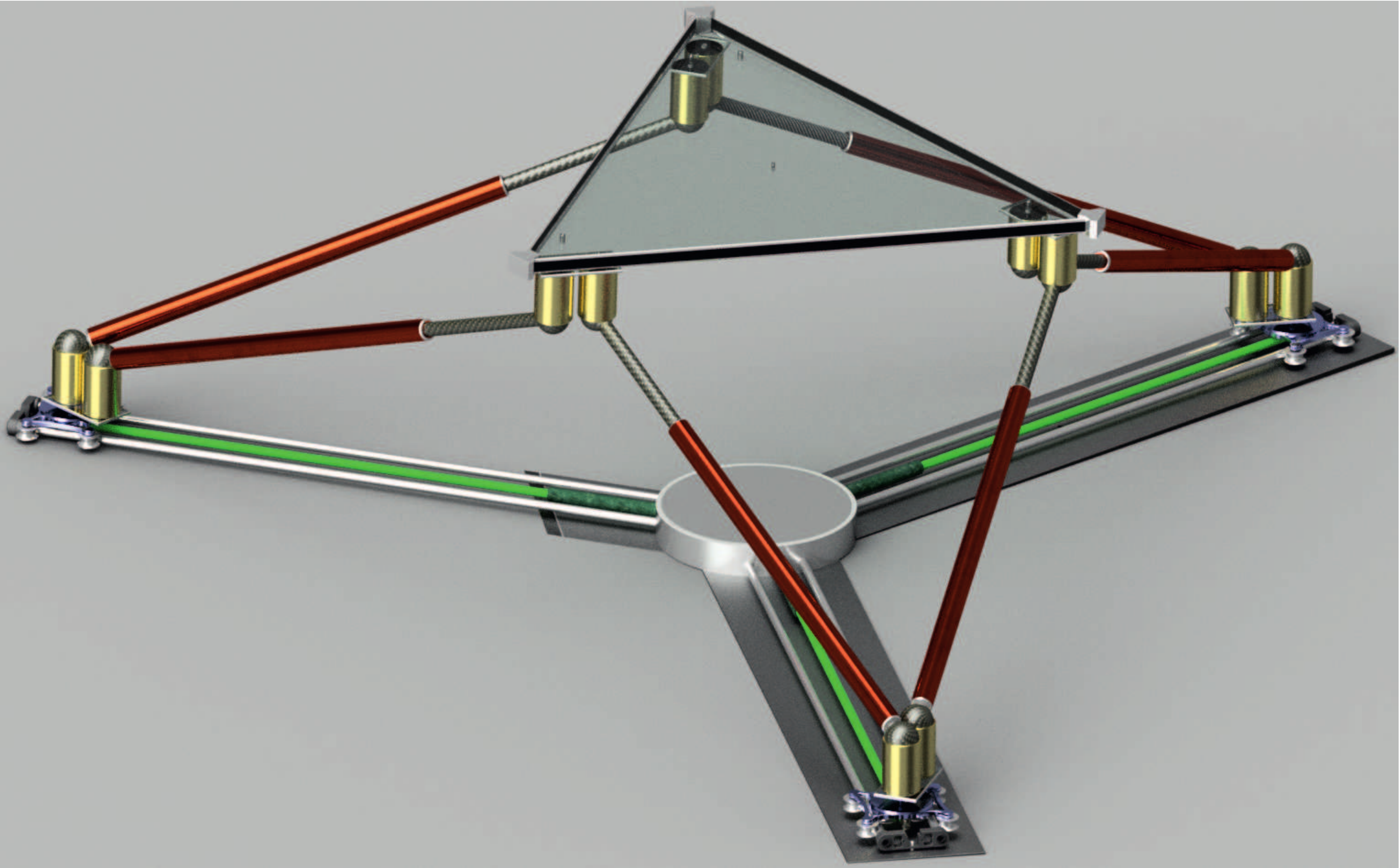}
  \end{overpic} 
 \begin{overpic}
    [height=42mm]{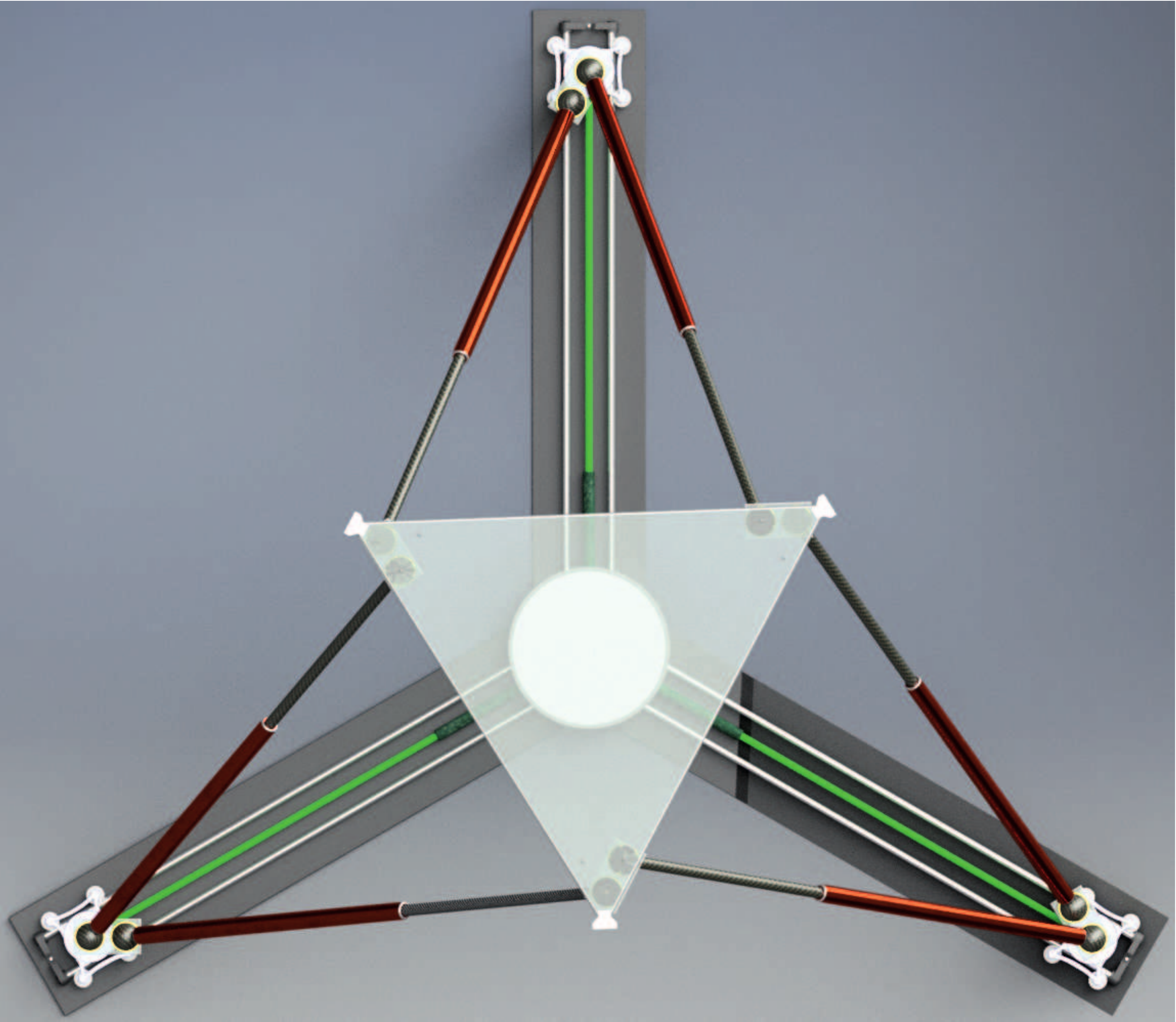}
  \end{overpic} 
\end{center} 
\caption{Axonomatic view (left) and top view (right) of a kinematic equivalent mechanism to Fig.\ref{fig2}. 
On this design further studies (e.g.\  construction of a small scale prototype, optimal reconfiguration 
of the base for a given motion, \ldots) are based.}
\label{fig2b}
\end{figure}

But we pursue another strategy for the technical realization; namely the usage of a design which is totally free of 
double joints but kinematically equivalent to the octahedral manipulator (cf.\ \cite{rojas}). 
Therefore the resulting design illustrated in Fig. \ref{fig2b} can be built with standard components; i.e.\ without expensive custom products.

\begin{acknowledgement}
The first author is supported by Grant No.~P~24927-N25 of the Austrian Science Fund FWF. 
The second author is funded by the Doctoral College "Computational Design" of TU Vienna. 
Moreover the authors want to thank Hannes Kaufmann  
of the "Center for Geometry and Computational Desgin" at TU Vienna for fruitful discussions on VR motion simulations. 
\end{acknowledgement}

\newpage

\section*{Appendix A}

As already mentioned in the first bullet of Subsection \ref{sec:unavoidable}, the authors of \cite{schreiber} 
stated in Section Section V(D) that for the octahedral manipulator with three 
redundant legs {\it "any Cartesian pose of the moving platform can be reached with a non-singular 
configuration of the mechanism".}\footnote{This is only true under Assumption 2  of \cite[Section V(D)]{schreiber}. 
Without this assumption all poses where the platform and the base are coplanar are "unavoidable singularities".} 
Moreover, it is "{\it conjectured that this result can be extended to 
mechanisms whose spherical joints on the platform do not coincide by pairs".} 
In this section we construct a counter example to the conjecture. 

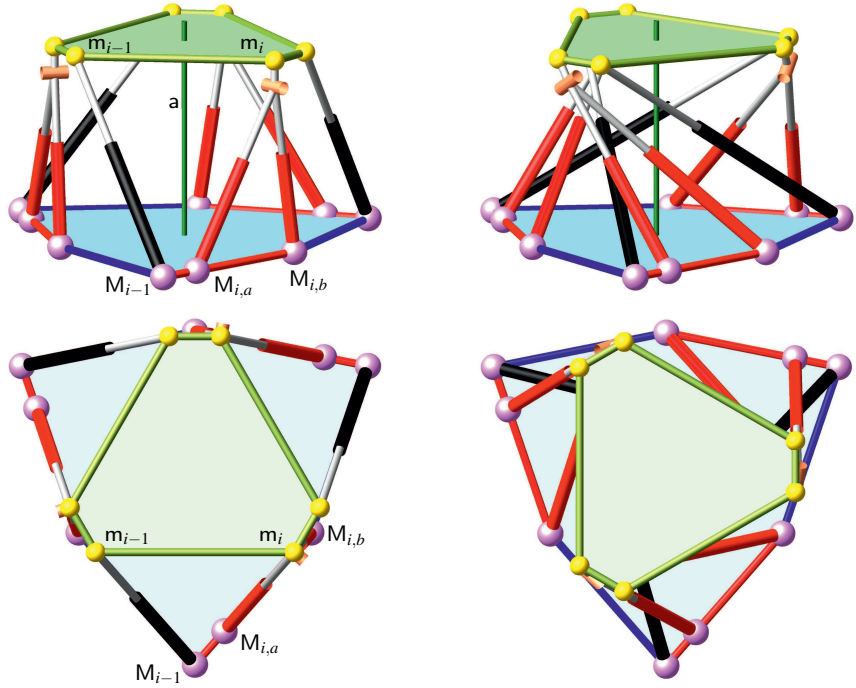
\begin{figure}[t!]
\begin{center} 
\begin{overpic}
    [width=50mm]{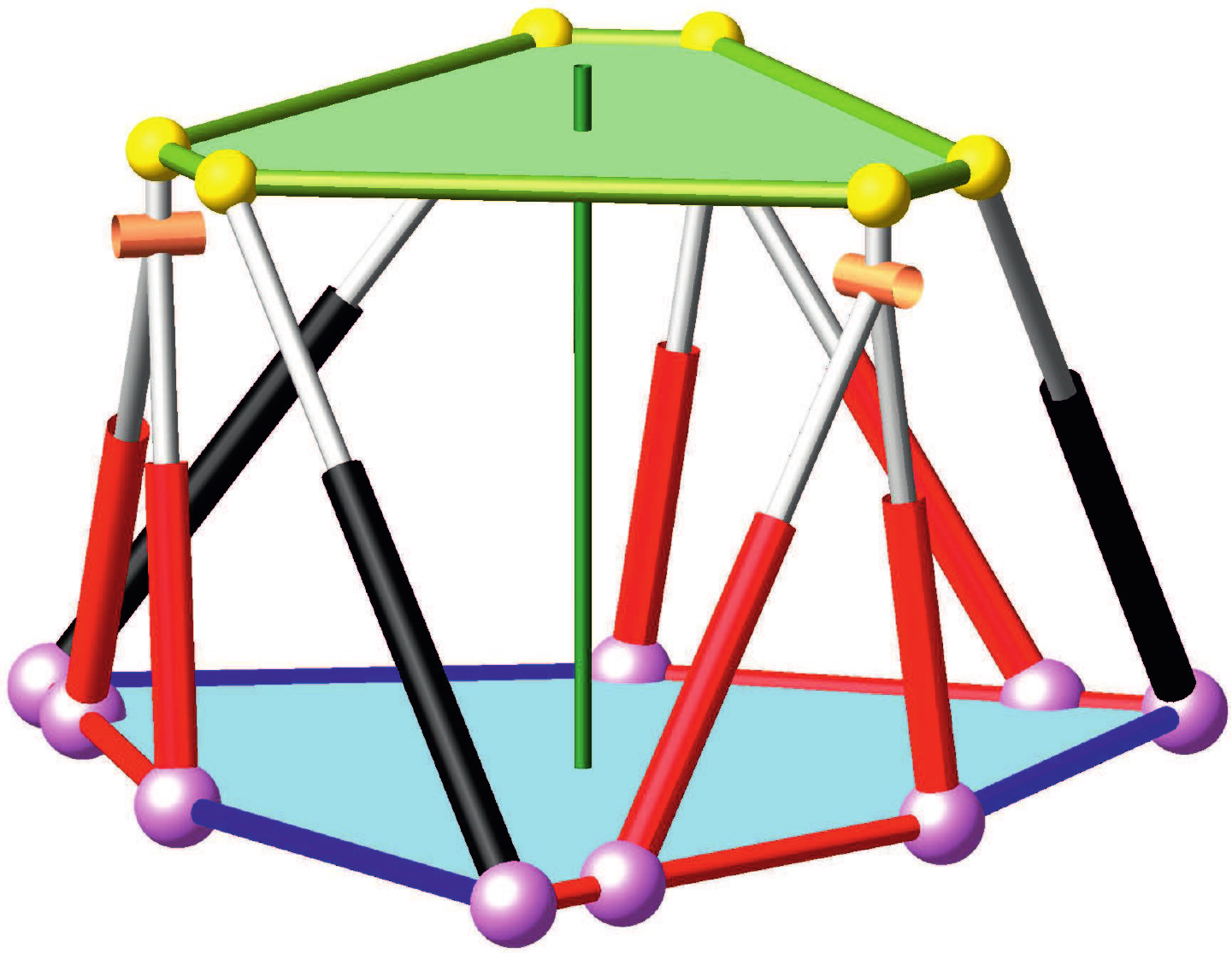}
\begin{small}
\put(43,49){$\go a$}
\put(26,0){$\go M_{i-1}$}
\put(55,0){$\go M_{i,a}$}
\put(75,1.5){$\go M_{i,b}$}
\put(21.5,64.5){$\go m_{i-1}$}
\put(62,64.5){$\go m_{i}$}
\end{small}     
  \end{overpic} 
\qquad \qquad
 \begin{overpic}
    [width=50mm]{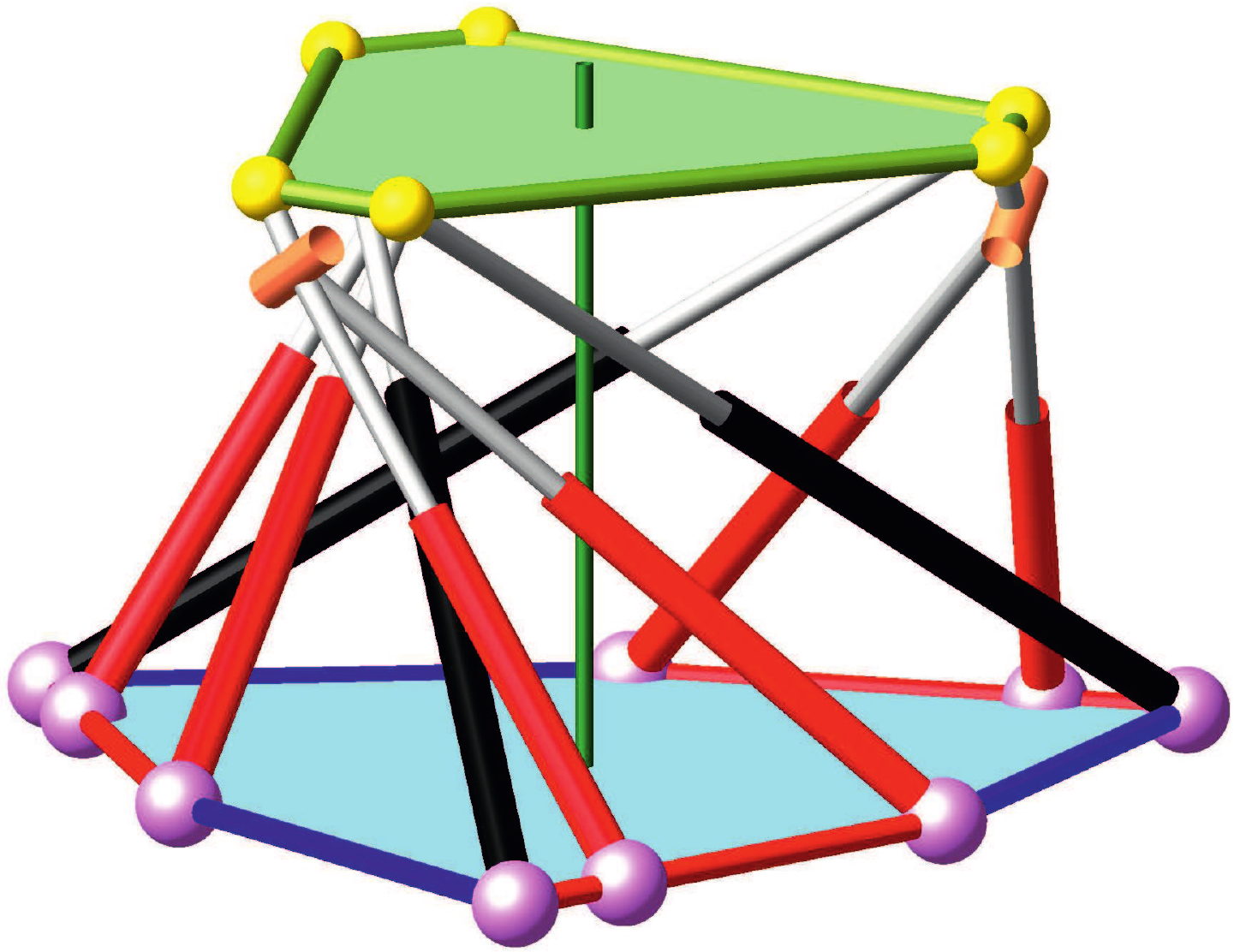}
\begin{small}
\end{small}         
  \end{overpic} 
\end{center} 
\begin{center} 
\begin{overpic}
    [width=50mm]{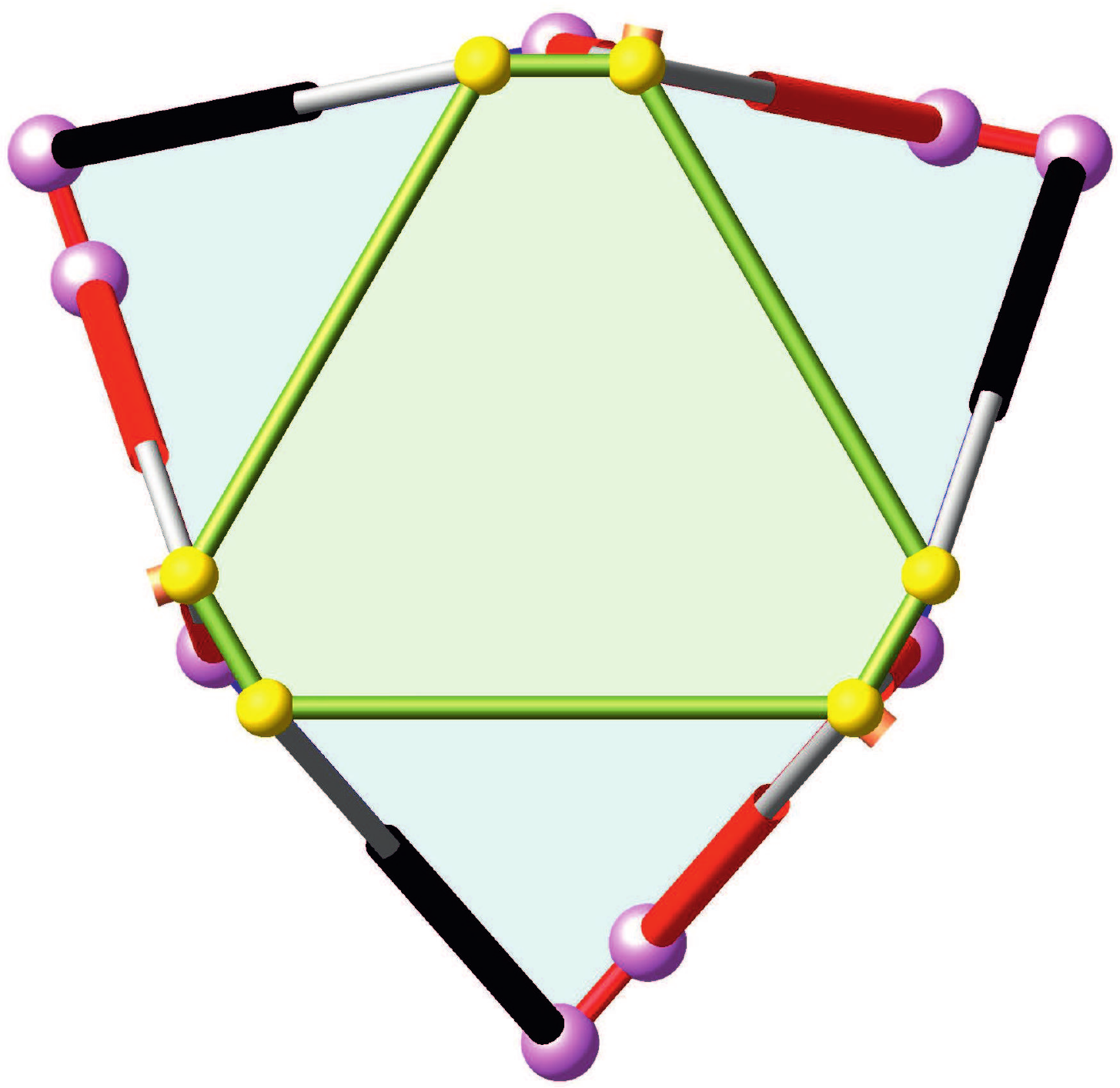}
\begin{small}
\put(34,0){$\go M_{i-1}$}
\put(62,8){$\go M_{i,a}$}
\put(85,37){$\go M_{i,b}$}
\put(26,37){$\go m_{i-1}$}
\put(67,37){$\go m_{i}$}
\end{small}     
  \end{overpic} 
\qquad \qquad
 \begin{overpic}
    [width=50mm]{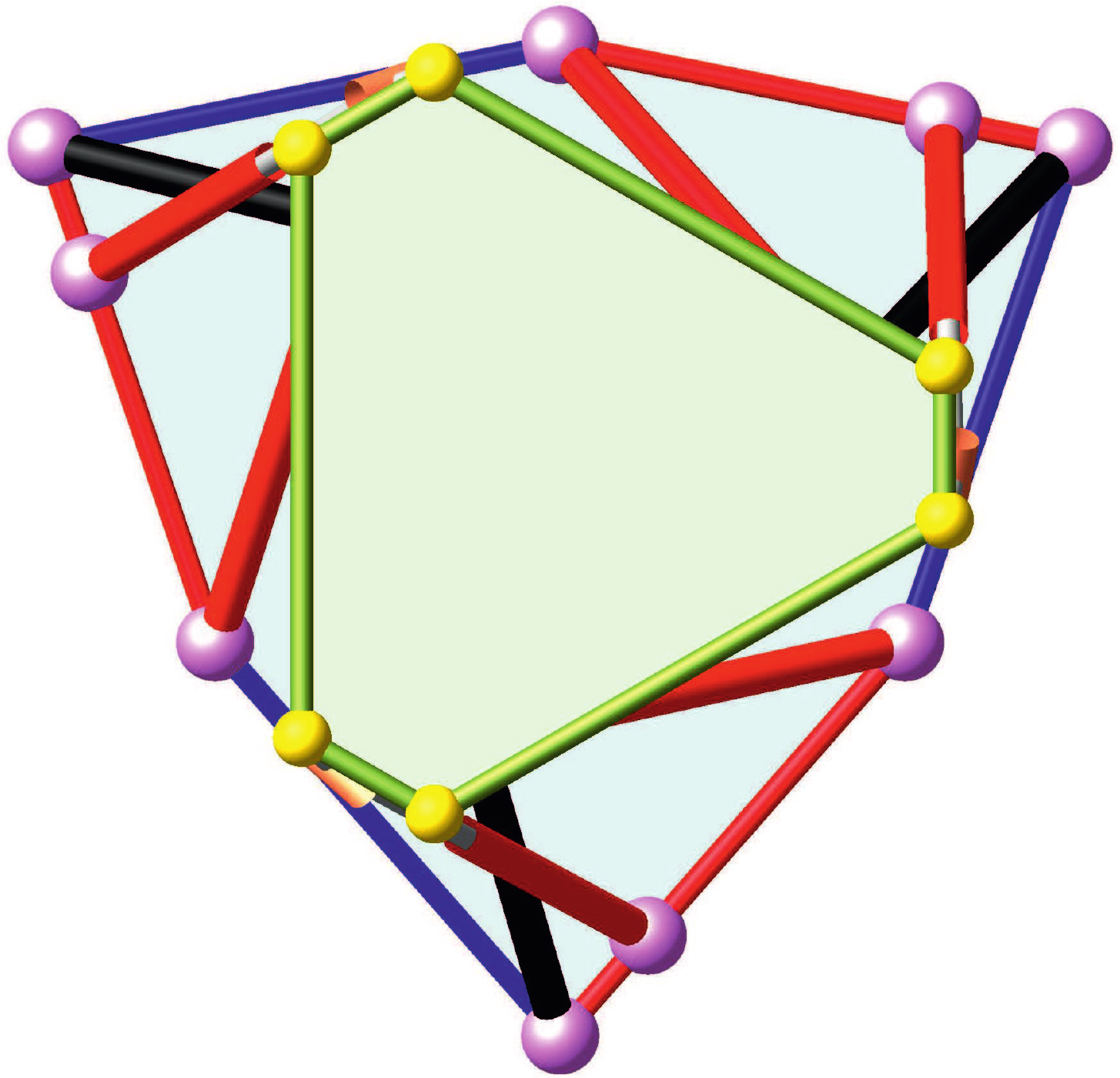}
\begin{small}
\end{small}         
  \end{overpic} 
\end{center} 
\caption{Non-redundant legs are displayed in black and redundant ones are illustrated in red. 
In the upper row the axonometric views and in the lower row the top views 
of a non-singular configuration (left) and an  unavoidable singularity (right), respectively, are given. 
}
\label{fig3}
\end{figure}

According to Assumption 1  of \cite[Section V(D)]{schreiber} the base anchor points have to fulfill the 
following property:
\begin{equation}\label{ass:1}
\go M_{i,a},\go M_{i,b},\go M_{i-1}\,\, \text{are collinear for $i=1,3,5$ (mod 6)},
\end{equation}
where $\go M_{i-1}$  is the base point of a non-redundant leg and  
$\go M_{i,a},\go M_{i,b}$ denote the two base anchor points of a redundant leg. 
The geometric structure of the redundant leg can be seen in Fig.\ \ref{fig3} (upper left). 
It should only be noted that the orange cylinders represent passive rotational joints and that  
the 3 gray bars linked by this joint are coplanar. 

We start with a semi-hexagonal platform parallel to the base plane $\delta$. Then we determine the corresponding base point(s) 
of $\go m_j$ as follows: We construct a plane $\varepsilon_j$ orthogonal to $\delta$, which  
touches the circumcircle of the semi-hexagonal platform in $\go m_j$. 
The intersection line of $\varepsilon_j$ and $\delta$ is illustrated in  Fig.\ \ref{fig3} (left) 
in red (for $j=1,3,5$) and blue (for $j=2,4,6$), respectively. 
Under consideration of (\ref{ass:1}) we select the base point(s) on these intersection lines.  

Then we rotate the platform by $-90$ degrees about the center axis $\go a$ and end up with a 
singular configuration (so-called Fichter singularity \cite{fichter}) illustrated in Fig.\ \ref{fig3} (right).  
It can easily be checked that this is an unavoidable singularity.

\section*{Appendix B}

We split the proof of Theorem \ref{thm1} into two subsections. In the first one very special cases are 
discussed and in the second one we attack the general  case.

\subsubsection*{Special cases}

\begin{enumerate}
\item
Within this first item we study all cases where at least two of the four Euler parameters are zero.
	\begin{enumerate}
	\item
	$e_1=e_2=0$: Then $Q$ and $L$ are fulfilled identically and we remain with $A_1A_2=0$, which factors into
	\begin{equation}
	z^3(e_0^2+e_3^2)(e_0-e_3)(e_0+e_3)=0.
	\end{equation}
	For $z=0$ the legs are in a field of lines, whose carrier plane is the base ($\Rightarrow$ row 1 of the table given in Theorem \ref{thm1}). 
	For $e_0=\pm e_3$ we get the Fichter singularity (cf.\ \cite{fichter}); i.e.\ rotation of the platform by $\pm 90^{\circ}$ about
	the z-axis ($\Rightarrow$ row 2).
	\item
	$e_0=e_3=0$: This is already a singular configuration ($\Rightarrow$ row 3), which 
	is not of practical relevance as the platform is upside down.
	\item
	$e_0=e_2=0$ and $e_1e_3\neq 0$: It can easily be seen that  $Q=L=A_1A_2=0$ can only hold for $z=e_1e_3$ ($\Rightarrow$ row 13). 
	The lines are in a singular linear line complex. 
	\item
	$e_1=e_3=0$ and $e_0e_2\neq 0$: Analogous considerations as in item (1c) yield $z=-e_0e_2$ ($\Rightarrow$ row 14). 
	The lines are in a singular linear line complex. 
	\item
	$e_0=e_1=0$ and $e_2e_3\neq 0$: It is not difficult to verify that the unavoidable singularities are given by 
	$y=z=0$ ($\Rightarrow$ row 15). 
	\item
	$e_2=e_3=0$ and $e_0e_1\neq 0$: Analogous considerations as in item (1e) yield
	$y=z=0$ ($\Rightarrow$ row 16). 
	\end{enumerate} 
\item
$e_0=e_3\neq 0$: In this case $Q$ factors into 
\begin{equation}
2e_3^3(e_1+e_2)(e_1-2e_2-\sqrt{3}e_2)(e_1-2e_2+\sqrt{3}e_2).
\end{equation}
We remain with the following three cases:
	\begin{enumerate} 
	\item
	$e_2=-e_1$ and $e_3\neq 0$: Now we get 
	\begin{equation}
	\begin{split}
	L&=4e_1^2e_3(z-2e_1e_3)(e_1z+2e_1^2e_3+2e_3^3+ye_3), \\
	A_1A_2&=4xe_1e_3(z-2e_1e_3)(2e_1e_3y-e_3^2z+e_1^2z),
	\end{split}
	\end{equation}
	thus the following cases can occur: 
		\begin{enumerate}
		\item
		$e_1=0$: We get again the Fichter singularity. 
		\item
		$z=2e_1e_3$ and $e_1\neq 0$: In this case all legs intersect a line and therefore they belong to 
		a singular linear line complex ($\Rightarrow$ row 4, upper sign).
		\item
		$e_1z+2e_1^2e_3+2e_3^3+ye_3=0$ and $(z-2e_1e_3)e_1\neq 0$: We can solve this equation for $z$ and then  
		$A_1A_2$ can only vanish without contradiction for  
		$x=0$ ($\Rightarrow$ row 5, upper sign) or  $y = 2e_1^2-2e_3^2$ ($\Rightarrow$ row 6, upper sign). 
		Therefore we get two lines in the position space.
		\end{enumerate}
	\item
	$e_1= (2+\sqrt{3})e_2$ and $(e_1+e_2)e_3\neq 0$: By inspecting $L$ and $A_1A_2$ in a similar way as 
	done in very detail in item (a) we get the following solutions: 
		\begin{enumerate}
		\item
		$z=-\frac{4e_2e_3}{\sqrt{3}-1}$: In this case all legs intersect again a line and therefore they belong to 
		a singular linear line complex ($\Rightarrow$ row 7, upper sign). 
		\item
		$x=\tfrac{16e_2^2+8e_2^2\sqrt{3}+y-4e_3^2}{\sqrt{3}}$ and $z=\frac{8e_2e_3}{\sqrt{3}-1}$: 
		For a given orientation we get a line in the position space ($\Rightarrow$ row 8, upper sign).
		\item
		$x=-y\sqrt{3}$ and 
		$z = \frac{2e_3(e_3^2+4e_2^2+2e_2^2\sqrt{3}-y)}{e_2(\sqrt{3}+1)}$: 
		For a given orientation we get again a line in the position space ($\Rightarrow$ row 9, upper sign).
		\end{enumerate}
	\item
	$e_1= (2-\sqrt{3})e_2$ and $(e_1+e_2)e_3\neq 0$: Exactly the same computations as in item (b) yield 
	the solutions given in the rows 7, 8, and 9, but now for the lower signs. 
\end{enumerate}
\item
$e_0=-e_3\neq 0$: This is the same discussion as in item (2). The resulting solutions are given in 
the rows 4, 5 and 6, but now for the lower signs, and in the 
rows 10, 11 and 12 (upper and lower  signs). 
\end{enumerate}

\subsubsection*{General  case}

We start by solving $Q=0$ for $z$ which yields Eq.\ \ref{coord:z}. 
The denominator vanishes only for the special cases (1a), (1b), (2) and (3), respectively. 
We insert this expression for $z$ into $L=0$ and $A_1A_2=0$ and consider the greatest common divisor of the 
resulting numerators, which reads as follows:
\begin{equation}
(e_0e_1+e_2e_3)(e_2^2e_3^2-3e_2^2e_0^2+8e_2e_1e_0e_3-3e_1^2e_3^2+e_1^2e_0^2).
\end{equation}
If one of these two factors vanish we get the solutions given in the rows 17 and 18, respectively.  
The remaining factor of $L$ has $40$ terms and is linear in $x$. As the coefficient of $x$ equals
\begin{equation}
2(e_2^2+e_1^2)(e_0-e_3)(e_0+e_3)(e_0e_1-e_2e_3)(e_0e_2+e_1e_3)
\end{equation}
we have to distinguish the following three cases:

\begin{figure}[t!]
\begin{center} 
\begin{overpic}
    [height=43mm]{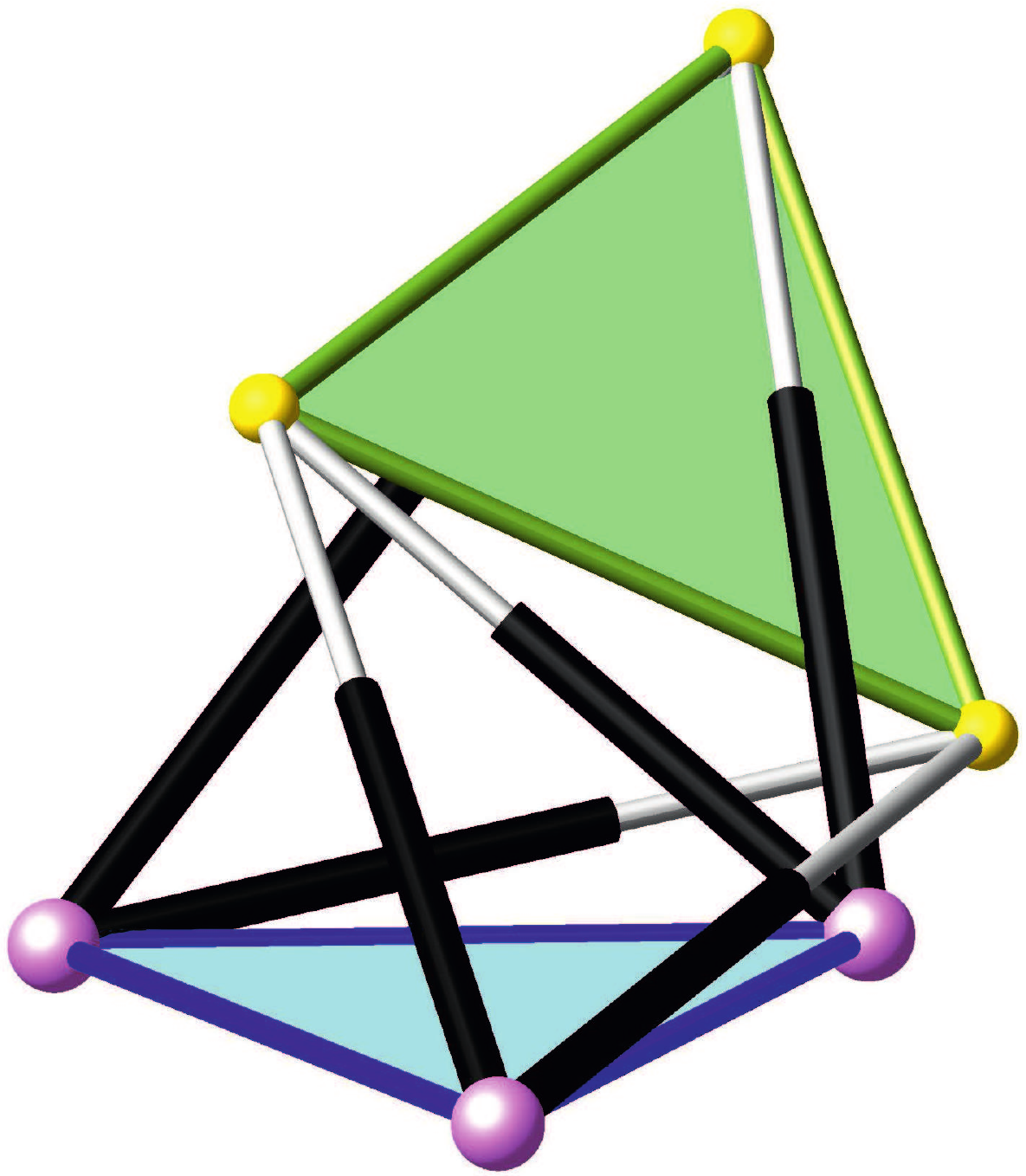}
  \end{overpic} 
\qquad \qquad 
 \begin{overpic}
    [height=40mm]{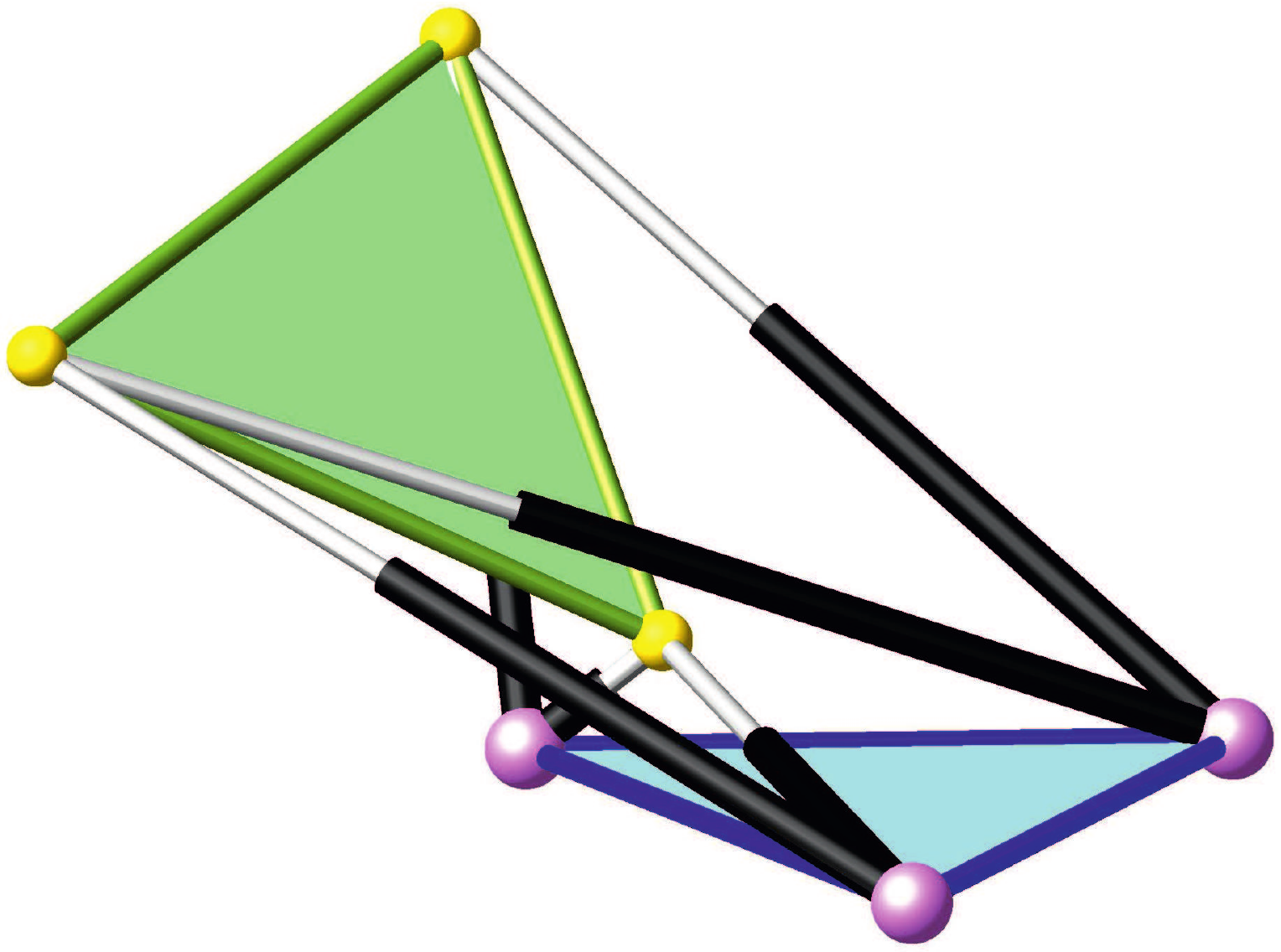}
  \end{overpic} 
\end{center} 
\caption{For the orientation given by the Euler parameters
$e_0=\frac{4\sqrt{105}}{175}$, $e_1=\frac{\sqrt{105}}{21}$,  $e_2=\frac{8\sqrt{105}}{105}$,  $e_3=-\frac{16\sqrt{105}}{525}$ 
the unavoidable singularities of row 21 and row 22 are illustrated in the left and right figure, respectively. 
The position of row 21 equals $x=-\frac{148327}{130830}$, $y=-\frac{66032}{65415}$, $z=\frac{12304}{13083}$
and the one of row 22 is given by $x=\frac{40969}{65415}$, $y=-\frac{85772}{65415}$, $z=\frac{12304}{13083}$. 
}
\label{fig4}
\end{figure}

\begin{enumerate}
\item
$e_0e_1-e_2e_3=0$: 
If one of the $e_i$'s is equal to zero, this equation implies that also a second one has to vanish. 
Therefore we get one of the already discussed special cases (1a,b,c,d). 
As a consequence we can set $e_0=\frac{e_2e_3}{e_1}$. Then the numerator of $L$ simplifies to
\begin{equation}
e_3^4(e_1-e_2)(e_1+e_2)(e_2^2+e_1^2)^3(2e_2e_1^2-ye_1+2e_2e_3^2).
\end{equation}
As $e_1=\pm e_2$ implies $e_0=\pm e_3$ we can only end up with a case studied in the special case (2) 
and (3), respectively. Therefore we only remain with $2e_2e_1^2-ye_1+2e_2e_3^2=0$ which implies 
row 19.
\item
$e_0e_2+e_1e_3=0$: For the same reasons as in item (1) of the general case 
we can set $e_0=-\frac{e_1e_3}{e_2}$. Analogous considerations imply the 
condition $2e_1e_2^2-2e_1e_3^2+ye_2=0$ resulting in row 20.
\item
$(e_0e_1-e_2e_3)(e_0e_2+e_1e_3)\neq 0$: In this case we can solve $L=0$ for $x$. 
Plugging the resulting expression into $A_1A_2$ yields two factors, which are linear in $y$. 
They imply the solutions given in row 21 and 22, which are illustrated in Fig.\ \ref{fig4}. 
\end{enumerate}
End of all cases. \hfill $\BewEnde$

\end{document}